\documentclass{INTERSPEECH2023}
\interspeechcameraready 

\usepackage{multirow}
\usepackage{siunitx}

\title{Inconsistency Ranking-based Noisy Label Detection for High-quality Data}
\name{Ruibin Yuan$^{*1,2}$, Hanzhi Yin$^{*1}$, Yi Wang$^1$, Yifan He$^1$, Yushi Ye$^1$, Lei Zhang$^{2}$, Zhizheng Wu$^{3}$}
\address{
    $^{1}$Carnegie Mellon University, Pittsburgh, PA 15213, USA\\
    $^{2}$Stardust.ai, Beijing, China\\
    $^{3}$The Chinese University of Hong Kong, Shenzhen, China
}
\email{
    \{ruibiny,hanzhiy,wangyi,yifanhe,yushiye\}@andrew.cmu.edu\\
    lei.zhang@stardust.ai\\
    wuzhizheng@cuhk.edu.cn
}

\begin{document}
\maketitle
\begin{abstract}
The success of deep learning requires high-quality annotated and massive data.
However, the size and the quality of a dataset are usually a trade-off in practice, as data collection and cleaning are expensive and time-consuming. In real-world applications, especially those using crowdsourcing datasets, it is important to exclude noisy labels. To address this, this paper proposes an automatic noisy label detection  (NLD) technique with inconsistency ranking for high-quality data. 
We apply this technique to the automatic speaker verification (ASV) task as a proof of concept. We investigate both inter-class and intra-class inconsistency ranking and compare several metric learning loss functions under different noise settings. Experimental results confirm that the proposed solution could increase both the efficient and effective cleaning of large-scale speaker recognition datasets.\footnote{\text{\{*\}} denotes equal contribution.}\footnote{
Code will be open-sourced on GitHub.
}.
\end{abstract}

\section{Introduction}
The success of Deep Neural Networks (DNNs) heavily relies on large-scale datasets with high-quality annotations.
However, in practice, datasets' size and quality are usually a trade-off,
as collecting reliable datasets is labor-intensive and time-consuming.
This implies that datasets usually contain some hard-to-estimate inaccurate annotations,
often referred to as ``noisy labels''.
Noisy labels may over-parameterize DNNs and lead to performance degradation due to the memorization effect.
These pose bottlenecks in training and employing DNNs in real-world scenarios.

Various strategies have been proposed to mitigate the negative effects of noisy labels, usually referred to as noisy-label learning (NLL) techniques \cite{zhang2020decoupling}.
The loss correction technique is one of the effective ways \cite{zhang2018generalized, patrini2017making, arazo2019unsupervised, wang2019symmetric, hendrycks2018using, lukasik2020does}.
It usually involves modifying the loss function to make it robust to the label noise or correcting the loss values according to the label noise.
Another approach is to let neural nets correct the labels during training.
The correction signal can be computed from an external neural structure \cite{jiang2018mentornet} or algorithm \cite{goldberger2016training}, or internal structure, i.e., a noise layer \cite{bekker2016training}.
Co-teaching is another way of label correction.
It usually involves training two networks simultaneously through dataset co-divide, label co-refinement, and co-guessing, in order to filter out label noise for each other \cite{li2020dividemix, han2018co, chen2020simple}.

The majority of such research studies are from the computer vision domain.
For the audio and speech domain, much less research on noisy labels has been conducted.
\cite{akiyama2019multitask} applied semi-supervised techniques to learn from soft pseudo labels and ensemble the predictions to overcome noisy data.
\cite{zheng2019towards} explored the extent of noisy labels’ influences on trained x-vector embeddings \cite{snyder2018x}.
They showed that mislabeled data could severely damage the performance of speaker verification systems on the NIST SRE 2016 dataset \cite{sadjadi20172016} and proposed regularization approaches to mitigate the damage.
Later, \cite{li2021speaker} also demonstrated that label noise leads to significant performance degradation for both the x-vector front-end and PLDA back-end on NIST SRE 2016 and proposed a few strategies to resist label noise. 
Contrary to previous results, \cite{pham2020does} experimented with simulating noise on the VoxCeleb2 dataset \cite{chung2018voxceleb2} and showed that even high levels of label noise had only a slight impact on the ASV task, using either GE2E or CE loss functions. However, the noises they investigate are all closed-set noise \cite{sachdeva2021evidentialmix}, and never consider another type of frequently occurring noise, open-set noise.
The most similar work to this paper is \cite{qin2021simple}, which proposed an iterative filtering method to remove noisy labels from the training set iteratively in order to improve the speaker recognition performance on the VoxCeleb dataset. However, it only focuses on mitigating the noise effect under a small amount of noise and does not investigate the effect of different noise settings and corresponding noisy label detection solutions. 

In this paper, instead of directly mitigating the effects of label noise, we try to address this problem from the root cause: we focus on the detection of noisy labels.
Detecting and filtering noisy labels benefits not only the model training but also the crowdsourcing data platforms,
e.g., Amazon Mechanical Turk and Scale AI. The filtered noisy labels can be used for relabeling to build higher quality datasets or identifying the noisy label providers, i.e., cheaters. 
Specifically, we investigate noisy label detection (NLD) under the context of speaker recognition, as this has been an under-explored topic. Yet, it plays an important role in collecting and correcting speaker identity annotations in large-scale speech datasets.
Our contributions are two folds:
\begin{itemize}
\item We are the first to comprehensively compare common metric learning loss functions used in the ASV task under various noise settings, like open-set and closed-set noise, with different noise levels, on a relatively large scale for the first time.
\item We propose two novel ranking-based NLD methods based on the inconsistent nature of the noisy labels and achieve over $90\%$ detection precision even under an extreme $75\%$ level of noise.
\end{itemize}

\section{Method}
\begin{figure}[th]
    \vspace{-1em}
    \setlength{\abovecaptionskip}{0.2cm}
    \setlength{\belowcaptionskip}{0.2cm}
    \centering
    \includegraphics[width=\columnwidth]{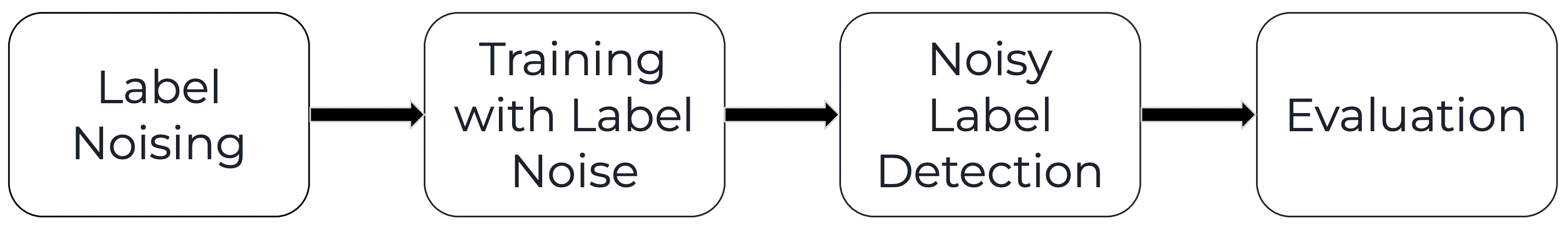}
    \caption{The pipeline.}
    \label{fig:pipeline}
    \vspace{-1.5em}
\end{figure}

Our pipeline is shown in Fig.~\ref{fig:pipeline}.
Firstly, we add noise to a clean dataset to simulate real-life noise.
Then, we train an X-vector with commonly used metric learning loss functions under different noise settings.
With the learned speaker embedder, we propose two inconsistency-ranking-based NLD methods.
Finally, we compare and evaluate the proposed methods.

\subsection{Label Noising} \label{section:method-noise}
Suppose we scraped a large-scale speaker recognition dataset
from the web, two types of label noise, 
the closed-set, and the open-set noise, may exist in the dataset. 
They can be simulated using a clean dataset $\mathcal{D}$
and an auxiliary dataset $\mathcal{D}'$ with no class overlap. 
Given the noise level $q\%$, the simulation can be done as follow.

\noindent\textbf{Close-set (permute) noise.}
For each utterance in $\mathcal{D}$,
we replace its true \textit{label} with another label
randomly drawn from $\mathcal{D}$ by a chance of $q\%$.
It simulates that some utterances
of an existing (in-distribution) speaker
are mistakenly assigned to another existing speaker class. 

\noindent\textbf{Open-set noise.}
For each utterance in $\mathcal{D}$,
we replace its \textit{utterance} with another utterance
randomly drawn from $\mathcal{D}'$ by a chance of $q\%$
with the label untouched.
It simulates some utterances from out-of-distribution speakers
are mistakenly assigned to an existing speaker class.

\subsection{Training Loss Functions} \label{section:loss-function}
We investigate the effectiveness of four commonly used
loss functions under noisy label settings.
Suppose a dataset contains $C$ speaker classes.
During training, each mini-batch contains ${bs}$ utterances
whose embeddings are $\mathbf{x}_i$,
and the corresponding labels are $y_i$,
where $i \in \{1, 2, \dots, {bs}\}$ and $y \in \{1, 2, \dots, C\}$.

\noindent\textbf{Softmax Loss (CE).}
Softmax loss is also known as softmax cross-entropy (CE) loss
in most classification tasks.
It uses a fully-connected layer (FC layer) that maps an embedding
to a probability distribution over all classes,
which is then used for computing the cross entropy.
Softmax loss has the advantage of fast and easy convergence,
but it does not explicitly enforce intra-class compactness
and inter-class separation.

\noindent\textbf{Additive Angular Margin Loss (AAM) \cite{deng2019arcface}.}
We can drop the bias and normalize the weight in the FC layer
and normalize the embedding vector $\mathbf{x}_i$.
The normalization scales everything onto a hypersphere.
By doing so, the matrix multiplication result of the FC layer
is the cosine of the angle between each weight vector
and the embedding vector, which is the cosine similarity.
We use $\cos\left(\varphi_{j, i}\right)$ to denote the cosine similarity
between the class $j$'s weight and the speaker $i$.
Now, the original softmax loss can be transformed
into a normalized softmax loss (NSL).
As the NSL only penalizes classification errors, 
embeddings learned by it are not sufficiently discriminative.
AAM resolves this by introducing two requirements.
First, AAM introduces an additive angular margin $m$
between each speaker class.
A normalized embedding vector can be classified as class $i$
only if $\cos(\varphi_{y_i, i} + m)$ is the largest
among all other cosine similarities without the margin.
Second, AAM scales all cosine similarities by a factor of $s$,
also known as the radius of the hypersphere,
to further diversify the target class and the negative class.

\noindent\textbf{Sub-center Additive Angular Margin Loss (AAMSC) \cite{deng2020subcenterarcface}.}
AAMSC introduces sub-classes to fortify AAM to combat noisy data.
In AAMSC, each speaker class has $K$ corresponding sub-classes.
When training, each embedding may take any sub-class from a speaker class,
For each sub-class cluster, the class with the highest cosine similarity
would be selected.
The rest procedures are the same as AAM.

\noindent\textbf{Generalized End-to-end Loss (GE2E) \cite{wan2018generalized}.}
GE2E is a contrastive learning loss.
It computes the self-excluded centroid from each batch,
maximizes the cosine similarities of utterance-centroid pairs within one class,
and minimizes the cosine similarities among different classes.
Contrary to previously mentioned losses,
which uses an FC layer as the parameterized classifier,
GE2E does not require any parameterized classifier.
The centroid calculation and embedding discrimination
are accomplished merely within a batch,
which requires speaker grouped batch sampling strategy.

\subsection{Noisy Label Detection by Inconsistency Ranking}
Clean training samples have consistent patterns,
while noisy ones are mostly inconsistent.
As neural nets are good pattern recognizers,
a speaker embedder can learn consistent patterns
even from a large amount of noise before memorizing noise.
We can leverage this learned consistency
and detect the noisy samples according to their inconsistency.
The higher the inconsistency of a sample,
the more likely the sample being mislabeled.
Therefore, we propose that noisy samples can be identified
by ranking the inter-class and intra-class inconsistency. 

\noindent\textbf{Intra-class Inconsistency.}
A speaker embedder can learn relatively compact
representations of the same speaker. 
We compute each speaker's embedding centroid
and rank the cosine distances
between every utterance and its class centroid.
Clean samples should have consistently low distances,
while noisy samples should have inconsistent distances,
which are usually much higher than clean ones.
We define this as intra-class inconsistency.
Given a speaker $p \in \{1, 2, \dots, C\}$ with $U_p$ utterances in total,
The speaker centroid is computed by:

\begin{equation}
\mathbf{c}_p=\frac{1}{U_p} \sum_{u=1}^{U_p} \mathbf{x}_{p, u}
\end{equation}

\noindent where $\mathbf{x}_{p, u}$ denotes the embedding vector, and $u$ denotes the utterance index.
The intra-class inconsistency of utterance ${x}_{p, u}$ is defined as:

\begin{equation}
\mathbf{I}_{d} \left( \mathbf{x}_{p, u}, \mathbf{c}_s \right)
= 1 - \dfrac{\mathbf{x}_{p, u} \cdot \mathbf{c}_p}{\left\Vert\mathbf{x}_{p, u}\right\Vert_{2} \cdot\left\Vert\mathbf{c}_p\right\Vert_{2}}
\end{equation}

\noindent\textbf{Inter-class Inconsistency.} \label{section:inter-class}
A classifier $\mathbf{CLS}(\cdot; \theta)$ with parameter $\theta$
trained along with the embedder can estimate
the confidence of whether an utterance belongs to a certain class.
Usually, the class with the highest confidence
is consistent with the one-hot label vector $\mathbf{y}$, 
and the confidence gap between the ground-truth class and other classes should be consistently large.
However, the classifier may fail to maximize the confidence of the labeled class,
resulting in a less peaked, multi-peak, or even overly smooth confidence distribution.
In other words, the classifier has failed to separate different classes.
Such inter-class inconsistency suggests that the input sample may have been mislabeled.
The inter-class inconsistency is defined as:

\begin{equation}
\mathbf{I}_c= \min (1 - \mathbf{y} * \mathbf{CLS}(\mathbf{x} ; \theta))
\end{equation}

As aforementioned, GE2E does not have a parameterized classifier.
We manually construct a classifier with the speaker centroids for GE2E.
Also, when predicting $\mathbf{I}_c$ with AAM and AAMSC,
the FC layer is directly applied
without considering the margin and the scale.

\section{Experiments}

\noindent\textbf{Data Processing.}
We use VoxCeleb 2 as the training set $\mathcal{D}$
and VoxCeleb 1 as the auxiliary set $\mathcal{D}'$. 
We also use the VoxCeleb 1 test set
for evaluating equal error rate (EER).
All datasets are pre-processed to Fbank features
with sample rate \SI{16}{\kHz},
Hanning window with size \SI{0.025}{\second},
hop length \SI{0.01}{\second},
Mel bins \num{40}, and time frames \num{160}.
Simulated noises are applied to the training set.
We consider three noise levels
$q\in\{20, 50, 75\}$ with the two noise types
mentioned in Section \ref{section:method-noise}.
We include an extreme \qty{75}{\percent} noise level
as it not only tests our method's robustness
but also simulates extreme situations that may occur
in large-scale internet-scraped datasets.
We also train on the clean training set ($q=0$) for comparison.
In total, there are \num{6} noisy training sets
and one clean training set.

\noindent\textbf{Model Details.}
Our model is a three-layer \num{768} units
LSTM-based X-Vector encoder with an FC layer
attached to map model's outputs into speaker embeddings with size \num{256}. 
We keep the model simple and small as we do not intend
to obtain the optimal EER. 
We simply aim to obtain decent performance on the NLD task,
while testing a large number of settings
with a reasonable amount of computing budget.
One can always train with more complex architectures and strategies
to obtain better EER or even better NLD results
with a higher computing budget.
We train multiple models with different loss heads
attached to the output embeddings, including CE, AAM, GE2E, and AAMSC.
We use the Adam optimizer with a fixed learning rate \num{1e-4}. 
All models are trained for \num{75000} steps. 
We apply a speaker sampling strategy as we random sample $N$ speakers
with $M$ utterances for each speaker to balance the speaker distribution. 

\noindent\textbf{Hyperparameter Search.}
We conduct hyperparameter search for different losses,
and only report the best parameter settings. 
Each loss is tested for an effective batch size of ${bs}\in\{128, 256\}$. 
For CE, AAM and AAMSC, we set $N = bs$ and $M = 1$. 
For GE2E, we set $N={bs}/M$, where $M\in\{4, 8, 16\}$. 
Following \cite{chung2020defence}, for both AAM and AAMSC,
we set $s\in\{15, 30\}$, $m\in\{0.1, 0.2\}$.
Additionally, for AAMSC, we set $K\in\{3, 10\}$. 
To stabilize the training of marginal losses,
we turn on the easy margin for the first $1/8$
of the total training steps.
We only report the best settings.

\noindent\textbf{NLD Details.}
For NLD, we calculate and rank dataset-level inconsistencies
and take top $q\%$ largest samples as our predicted noisy label.
The prediction is then compared with the ground truth
to compute the precision score. 
During deployment, the prior knowledge of the noise level $q$ is not given.
This may be estimated by manual sampling
and labeling a small portion of data from the dataset.

\noindent\textbf{Reproducibility.}
To ensure the reproducibility of the experiment and reduce randomness,
we fix the random seed for each experiment and
repeat all experiments twice with seed \num{0} and seed \num{2},
reporting the average as the result.
All the experiments are run on a single Nvidia RTX 3090 GPU.
Training one model takes around \num{90} minutes, 
and finishing one round of experiments
with one random seed may take up to \num{2} weeks.
The code is released on GitHub. 
\begin{table}[h]
\setlength{\abovecaptionskip}{0.1cm}
\setlength{\belowcaptionskip}{-0.1cm}
\centering
\caption{Baseline EERs on Vox 1 Test. One can see the AAMSC achieves the best performance under most settings. Note that these systems are not trained to obtain the optimal EER. Instead, we train them for a reasonable and fixed amount of computing budget, sufficient to obtain decent performance on NLD tasks.}
\label{table:eer}
\scalebox{0.85}{
\begin{tabular}{|c|c|c|c|}
\hline
                       & Noise Level & Permute EER (\%)         & Open EER (\%)            \\ \hline
\multirow{4}{*}{CE}    & 0\%         & 9.56                     & 9.56                     \\
                       & 20\%        & 13.09                    & 25.96                     \\
                       & 50\%        & 25.00                    & 39.12                     \\
                       & 75\%        & 33.73                    & 44.28                     \\ \hline
\multirow{4}{*}{GE2E}  & 0\%         &  8.52                    & 8.52                     \\
                       & 20\%        & 10.40                    & 11.34                     \\
                       & 50\%        & \textbf{11.28}           & 14.38                     \\
                       & 75\%        & 34.51                    & 33.64                     \\ \hline
\multirow{4}{*}{AAM}   & 0\%         &  8.89                    & 8.89                      \\
                       & 20\%        &  9.65                    & 11.58                     \\
                       & 50\%        & 14.02                    & 16.92                     \\
                       & 75\%        & \textbf{24.38}           & 29.11                     \\ \hline
\multirow{4}{*}{AAMSC} & 0\%         & \textbf{7.33}            & \textbf{7.33}             \\
                       & 20\%        & \textbf{8.56}            & \textbf{8.55}             \\
                       & 50\%        & 12.91                    & \textbf{13.51}            \\
                       & 75\%        & 24.50                    & \textbf{26.20}            \\ \hline
\end{tabular}
}
\vspace{-1em}
\end{table}

\section{Result and Discussion}
\subsection{Comparison of loss functions under different noises}
Table \ref{table:eer} lists the best EERs
after hyperparameter search for each combination of
noise level, noise type, and loss function.
During the search, we found that a larger batch size is helpful to counter label noise.
Increasing $M$ of GE2E when the noise level goes up
and using larger $K$ for AAMSC are also helpful.
Comparing the performance of loss functions under different noise settings,
we have the following findings:

\noindent\textbf{(1) All loss functions exhibit varying EER increase
when noise is added to the dataset.}

\noindent\textbf{(2) Open-set noise is more harmful than close-set noise.}
We observe a performance drop on all loss functions
when comparing close-set noise and open-set noise under the same noise level.
Since open-set noise is very common in web-crawled datasets,
the community should pay more attention to it.

\noindent\textbf{(3) AAMSC is a more robust loss function under noise.} 
Even though the convergence of AAMSC is sometimes fragile,
it still achieves the best EER in most cases after some tuning.

\noindent\textbf{(4) CE's EER is more susceptible under increasing noise level.}
This may indicate that the learned embedding space
is susceptible to noise when using naive classification loss.

\noindent\textbf{(5) In our settings, GE2E does not perform well
under extreme noise level.}
As GE2E does not have parameterized class centroids,
the quality of the in-batch class centroids heavily depends on the batch size. 
Since our batch size is fixed to control the computing budget,
$N$ and $M$ cannot be large simultaneously,
harming the in-batch centroid quality.
Nevertheless, given the success of weakly supervised contrastive methods
in vision-language \cite{radford2021learning},
we believe that with a higher budget and a significantly larger batch size,
GE2E may still have the potential to counter large label noise. 
Furthermore, unlike marginal losses, GE2E does not suffer from convergence issues. 
However, testing this speculation is beyond our scope and budget. 

\noindent\textbf{(6) AAM outperforms GE2E under extreme noise levels
but achieves similar or slightly worse EER under other noise settings.}

\subsection{Comparison of NLD methods}
Table \ref{table:intra-class} and \ref{table:inter-class} listed all NLD precisions
with the two proposed inconsistency ranking methods.
We have the following findings.

\noindent\textbf{(1) EER and NLD precision are not strongly aligned.}
Systems without state-of-the-art EER or even with a bad EER still perform well on the NLD task.
As long as the model can learn to extract consistent patterns of clean samples, the noisy labels can be well-detected with our proposed inconsistency ranking.

\noindent\textbf{(2) The inter-class method is more robust than the intra-class.} 
All loss functions reach useful NLD precision (close to or over \qty{90}{\percent})
with the inter-class method under all noise settings,
including the extreme \qty{75}{\percent} noise level. 
In many \qty{75}{\percent} noise level cases, we see a large improvement
when using the inter-class method compared to the intra-class method. 

\noindent\textbf{(3) The intra-class method performs similarly or slightly better
than inter-class in some less extreme noise cases.}

\noindent\textbf{(4) AAMSC achieves the best or close-to-the-best precision in most cases.} 
It works the best with the intra-class method and is close to best with the inter-class method.
This indicates that introducing noise tolerance sub-centers is effective.

\noindent\textbf{(5) Interestingly, CE performs the best with the inter-class method
but significantly worse with the intra-class method.} 
We suspect that the learned embedding space may be disturbed by noise,
and cosine distance fails in this space. 
However, the confidence calculation is still effective and robust. 

\noindent\textbf{(6) GE2E and AAM achieve the best results in some non-extreme noise settings.}
\begin{table}[h]
\setlength{\abovecaptionskip}{0.1cm}
\setlength{\belowcaptionskip}{-0.1cm}
\centering
\caption{Intra-class Inconsist. Ranking NLD Precision}
\label{table:intra-class}
\scalebox{0.9}{
\begin{tabular}{|l|lll|lll|}
\hline
\multirow{2}{*}{} & \multicolumn{3}{c|}{Permute Noise Lvl. (\%)}                             & \multicolumn{3}{c|}{Open Noise Lvl. (\%)}                       \\ \cline{2-7} 
 &
  \multicolumn{1}{c|}{20} &
  \multicolumn{1}{c|}{50} &
  \multicolumn{1}{c|}{75} &
  \multicolumn{1}{c|}{20} &
  \multicolumn{1}{c|}{50} &
  \multicolumn{1}{c|}{75} \\ \hline
CE                & \multicolumn{1}{l|}{88.12} & \multicolumn{1}{l|}{76.84}          & 79.61 & \multicolumn{1}{l|}{39.60} & \multicolumn{1}{l|}{24.50} & 70.39 \\ \hline
GE2E              & \multicolumn{1}{l|}{92.74} & \multicolumn{1}{l|}{\textbf{95.05}} & 79.75 & \multicolumn{1}{l|}{93.37} & \multicolumn{1}{l|}{95.64} & 76.16 \\ \hline
AAM               & \multicolumn{1}{l|}{92.98} & \multicolumn{1}{l|}{93.24}          & 80.63 & \multicolumn{1}{l|}{93.76} & \multicolumn{1}{l|}{94.92} & 83.68 \\ \hline
AAMSC &
  \multicolumn{1}{l|}{\textbf{93.71}} &
  \multicolumn{1}{l|}{93.13} &
  \textbf{81.00} &
  \multicolumn{1}{l|}{\textbf{94.79}} &
  \multicolumn{1}{l|}{\textbf{96.09}} &
  \textbf{89.17} \\ \hline
\end{tabular}
}
\end{table}
\begin{table}[ht]
\setlength{\abovecaptionskip}{0.1cm}
\setlength{\belowcaptionskip}{-0.3cm}
\centering
\caption{Inter-class Inconsist. Ranking NLD Precision}
\label{table:inter-class}
\scalebox{0.9}{
\begin{tabular}{|l|lll|lll|}
\hline
\multirow{2}{*}{} &
  \multicolumn{3}{c|}{Permute Noise Lvl. (\%)} &
  \multicolumn{3}{c|}{Open Noise Lvl. (\%)} \\ \cline{2-7} 
 & \multicolumn{1}{c|}{20} & \multicolumn{1}{c|}{50} & \multicolumn{1}{c|}{75} & \multicolumn{1}{c|}{20} & \multicolumn{1}{c|}{50} & \multicolumn{1}{c|}{75} \\ \hline
CE &
  \multicolumn{1}{l|}{91.37} &
  \multicolumn{1}{l|}{93.32} &
  \textbf{89.90} &
  \multicolumn{1}{l|}{91.39} &
  \multicolumn{1}{l|}{94.59} &
  \textbf{94.38} \\ \hline
GE2E &
  \multicolumn{1}{l|}{\textbf{92.93}} &
  \multicolumn{1}{l|}{\textbf{95.09}} &
  88.01 &
  \multicolumn{1}{l|}{90.27} &
  \multicolumn{1}{l|}{94.40} &
  89.44 \\ \hline
AAM &
  \multicolumn{1}{l|}{92.80} &
  \multicolumn{1}{l|}{92.96} &
  88.25 &
  \multicolumn{1}{l|}{\textbf{93.73}} &
  \multicolumn{1}{l|}{\textbf{95.37}} &
  93.57 \\ \hline
AAMSC &
  \multicolumn{1}{l|}{91.42} &
  \multicolumn{1}{l|}{92.74} &
  88.33 &
  \multicolumn{1}{l|}{92.43} &
  \multicolumn{1}{l|}{94.39} &
  93.43 \\ \hline
\end{tabular}
}
\vspace{-1em}
\end{table}

\subsection{Visualization of inconsistencies}
\begin{figure}[h]
\vspace{-1em}
\setlength{\abovecaptionskip}{0.1cm}
\setlength{\belowcaptionskip}{-0.3cm}
\begin{center}
    \centering
    \includegraphics[width=0.45\textwidth]{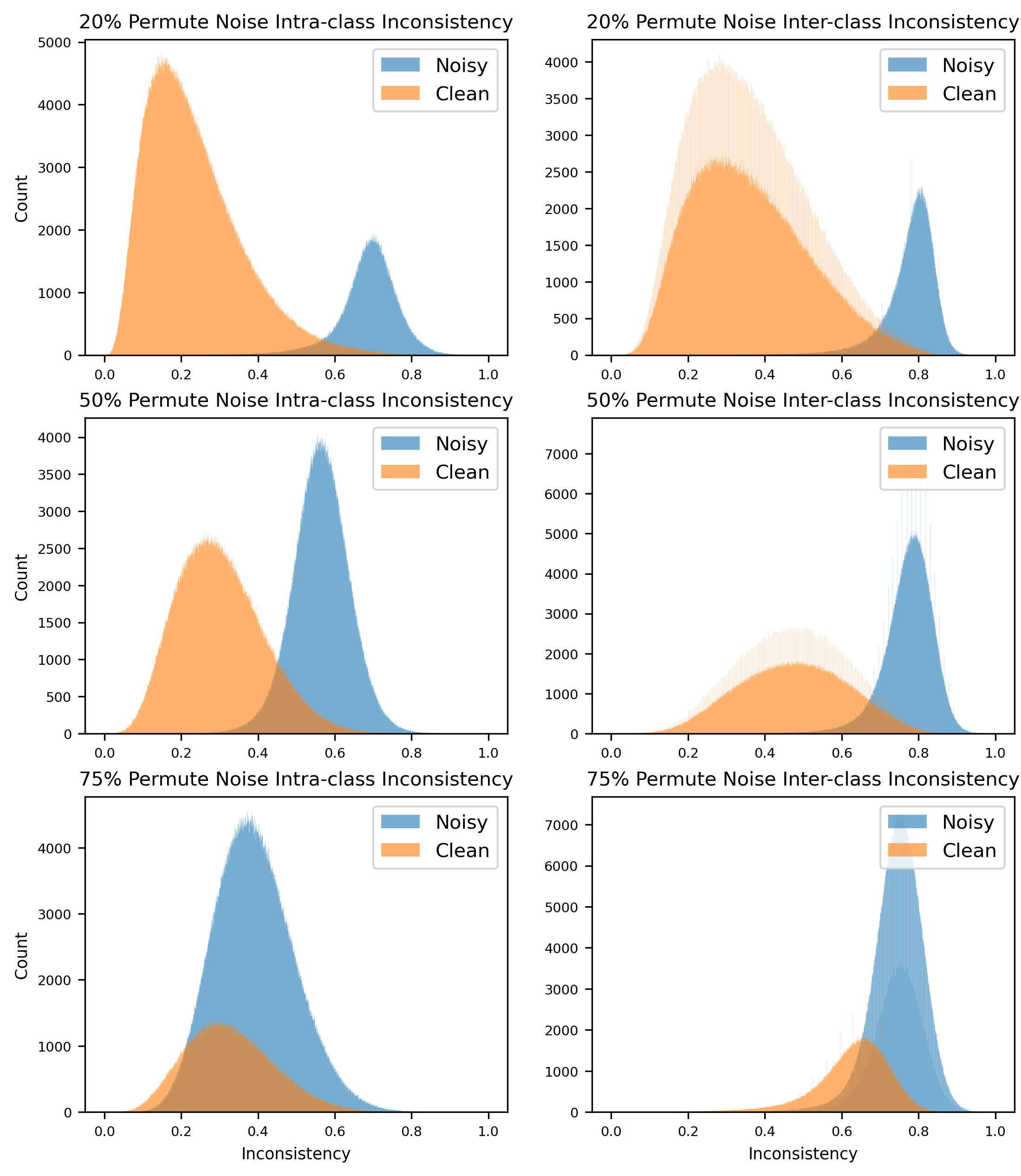}
    \caption{Visualizing Permute Noise Inconsistencies.}
    \label{fig:aamsc-inconsistency}
\end{center}
\vspace{-1em}
\end{figure}
In Fig.~\ref{fig:aamsc-inconsistency}, we plot
the normalized inconsistency distributions of Permute noise. 
We do not include the Open-set noise plots as they look similar to Fig.~\ref{fig:aamsc-inconsistency},
except that the clean and noise distributions are more separable in \qty{75}{\percent} noise level. 
We can see the boundary between clean and noise distribution in most cases,
except for the Permute \qty{75}{\percent} case. 
Also, the inter-class method has a clearer boundary than the intra-class method. 
Finally, the distributions become less separable as the noise level increases. 
But the overall good separation of clean and noise distribution explains
the good NLD performance in Table \ref{table:intra-class} and \ref{table:inter-class}.

\subsection{Retraining after NLD}
We then retrain our systems with AAMSC after removing the detected noisy samples with the proposed method to see if the EER improves. We use the same best settings from Table \ref{table:eer}. The result is shown in \ref{table:retraining}. We see significant EER reduction in almost all the settings except for Permute \qty{75}{\percent} intra-class. Especially in Open \qty{75}{\percent} inter-class, we can see the EER has almost cut in half. For the only unsatisfying result of Permute \qty{75}{\percent} intra-class, we believe that it is caused by the less robustness of the intra-class method and the fragility of the AAMSC, given previous analysis and the less separable boundary in Fig.~\ref{fig:aamsc-inconsistency}.
\begin{table}[h]
\setlength{\abovecaptionskip}{0.1cm}
\setlength{\belowcaptionskip}{-0.1cm}
\centering
\caption{
Retraining EERs of AAMSC after Removing Detected Noisy Samples with the Two Proposed Methods.
The baselines are the same AAMSC settings from Table \ref{table:eer}.
}
\label{table:retraining}
\begin{tabular}{|c|c|c|c|}
\hline
\multirow{2}{*}{Noise Level} & \multicolumn{3}{c|}{EER (\%)}              \\ \cline{2-4} 
                             & Baseline & Intra-class   & Inter-class    \\ \hline
Permute 20 \%                &  8.56    & \textbf{7.56} &  8.16           \\ \hline
Permute 50 \%                & 12.91    & \textbf{9.48} & \textbf{9.48}   \\ \hline
Permute 75 \%                & 24.50    & 36.60         & \textbf{17.65} \\ \hline
Open 20 \%                   &  8.55    & \textbf{7.64} &  7.80           \\ \hline
Open 50 \%                   & 13.51    & \textbf{8.80} &  9.25           \\ \hline
Open 75 \%                   & 26.20    & 18.82         & \textbf{15.03} \\ \hline
\end{tabular}
\vspace{-1em}
\end{table}

\section{Conclusion}
In this paper, we proposed two effective inconsistency ranking-based methods for automatic noisy label detection (NLD): intra-class and inter-class inconsistency ranking. 
We demonstrate that in the speaker verification task, the latter is more robust and achieves close to or over \qty{90}{\percent} NLD precision even under extreme \qty{75}{\percent} noise level.
We also extensively reviewed how noise types, noise levels, and loss functions affect training performance.
We found that open-set noise is more harmful than close-set noise,
and AAMSC loss achieves the best EER under most settings.
Surprisingly, a naive CE loss also performs fairly well in NLD
with the inter-class ranking method, despite its poor EER.
Overall, our findings suggest that given a classification dataset with unknown label noise,
the general solution to effectively detecting and cleaning noisy labels might be using the AAMSC loss with inter-class inconsistency ranking.

\section{Future Work}
Despite the promising performance of the proposed method, how to automatically estimate the noise level and determine the decision boundary for separating clean and noise distribution remains an open question.
Also, the effect of mixed noise is never investigated.
Future work may include using mixture models to estimate the noise level,
conducting more experiments on different mixed noise or new noise types.

\noindent\textbf{Acknowledgement}
The content is redacted for anonymity. 


\bibliographystyle{IEEEtran}
\bibliography{mybib}
\end{document}